\documentclass{article}

\usepackage[preprint]{neurips_2024}

\usepackage{amsmath}
\usepackage[utf8]{inputenc} 
\usepackage[T1]{fontenc}    
\usepackage{hyperref}       
\usepackage{url}            
\usepackage{booktabs}       
\usepackage{amsfonts}       
\usepackage{nicefrac}       
\usepackage{microtype}      
\usepackage{xcolor}         
\usepackage{graphicx}
\usepackage{cleveref}
\usepackage{cleveref}
\usepackage{amssymb}
\usepackage{wrapfig}
\usepackage{amsmath}
\usepackage{algorithm}
\usepackage{algorithmicx}
\usepackage{algpseudocode}
\usepackage{multirow}
\usepackage{latexsym}
\usepackage{diagbox}

\title{Segment Any 4D Gaussians}

\author{%
  Shengxiang Ji$^{1}$\footnotemark[1], \quad Guanjun Wu$^{1}$\footnotemark[1], \quad Jiemin Fang$^{2}$, \quad Jiazhong Cen$^{3}$, \and \textbf{Taoran Yi}$^{4}$, \quad  \textbf{Wenyu Liu}$^{4}$, \quad  \textbf{Qi Tian}$^{2}$, \quad \textbf{Xinggang Wang}$^{4}$\footnotemark[2] \\
  $^1$School of CS, Huazhong University of Science and Technology \;\;
  $^2$Huawei Inc. \\
  $^3$MoE Key Lab of Artificial Intelligence, AI Institute, Shanghai Jiao Tong University \\
  $^4$School of EIC, Huazhong University of Science and Technology\\
\texttt{\small \{jishengxiangzs,jaminfong\}@gmail.com} \\
\texttt{\small \{guajuwu,taoranyi,liuwy,xgwang\}@hust.edu.cn} \\
\texttt{\small jiazhongcen@sjtu.edu.cn,tian.qi1@huawei.com} \\
}

\begin{document}

\makeatletter
\g@addto@macro\@maketitle{
\begin{figure}[H]
    \setlength{\linewidth}{\textwidth}
    \setlength{\hsize}{\textwidth}
    \vspace{-7mm}
    \centering
    \includegraphics[width=1\linewidth]{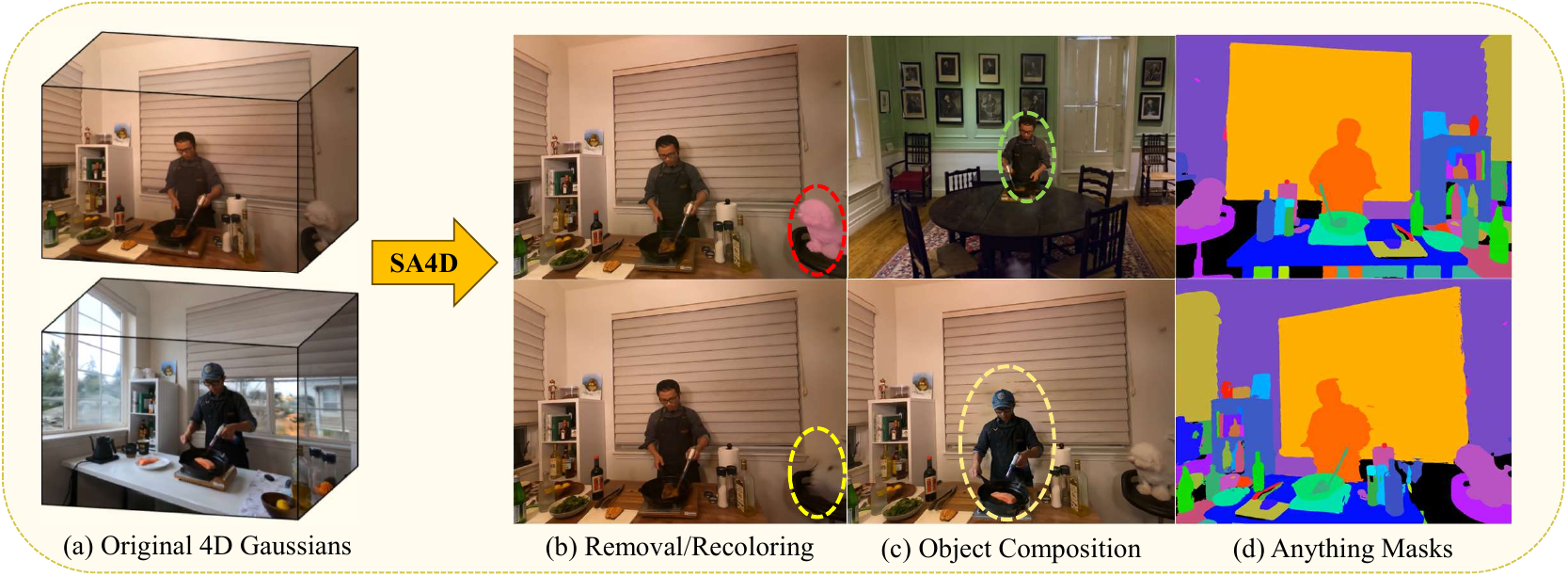}
  \vspace{-6pt}
  \caption{Applications of SA4D. Given pre-trained 4D Gaussians (column a), SA4D decomposes the Gaussians into the object level and supports object removal/recoloring (column b), composition (column c), and rendering anything masks (column d).}

    \label{fig:teaserfig}
\end{figure}
}
\makeatother  
\maketitle

\footnotetext[1]{Equal contributions.}
\footnotetext[2]{Corresponding author.}


\begin{abstract}
Modeling, understanding, and reconstructing the real world are crucial in XR/VR. Recently, 3D Gaussian Splatting (3D-GS) methods have shown remarkable success in modeling and understanding 3D scenes. Similarly, various 4D representations have demonstrated the ability to capture the dynamics of the 4D world. However, there is a dearth of research focusing on segmentation within 4D representations. 
In this paper, we propose Segment Any 4D Gaussians (SA4D), one of the first frameworks to segment anything in the 4D digital world based on 4D Gaussians. In SA4D, an efficient temporal identity feature field is introduced to handle Gaussian drifting, with the potential to learn precise identity features from noisy and sparse input. Additionally, a 4D segmentation refinement process is proposed to remove artifacts. Our SA4D achieves precise, high-quality segmentation within seconds in 4D Gaussians and shows the ability to remove, recolor, compose, and render high-quality anything masks. More demos are available
at: ~\href{https://jsxzs.github.io/sa4d/}{https://jsxzs.github.io/sa4d/}.
\end{abstract}



\section{Introduction}
\label{sec:intro}


Recently, SAM~\cite{kirillov2023segmentanything} has achieved great success in understanding the 2D image, and \cite{cen2023segment} extends it to the 3D representations. 3D Gaussian Splatting (3D-GS)~\cite{3dgs} has proven to be an efficient approach to modeling 3D scenes, which boosts the booming of 3D understanding~\cite{cen2023segmentany3dgs,zhou2023feature,ye2023gaussiangrouping}. However, our world lives in 4D. Therefore, achieving interactive and high-quality segmentation in 4D scenes remains an important and challenging task. We believe that 4D segmentation should achieve two goals: (a) Fast and interactive segmentation across the entire scene and (b) High-quality segmentation results. Some prior works~\cite{Dynamicviewsynthesisfromdynamicmonocularvideo,tian2023mononerf,robustdynamicradiancefields,jiang20234deditor} attempt to use binary segmentation masks to separate objects or train dynamic parts individually. However, they struggle to provide interactive and fast object-level segmentation masks. Meanwhile, directly incorporating segmentation features into 3D Gaussians~\cite{4dgs-2,chen2023periodic,yan2024street} can only address 6-DoF motion, such as cars, and may fall short in capturing non-rigid dynamics.



\begin{figure}[t]
    \centering
    \includegraphics[width=1\linewidth]{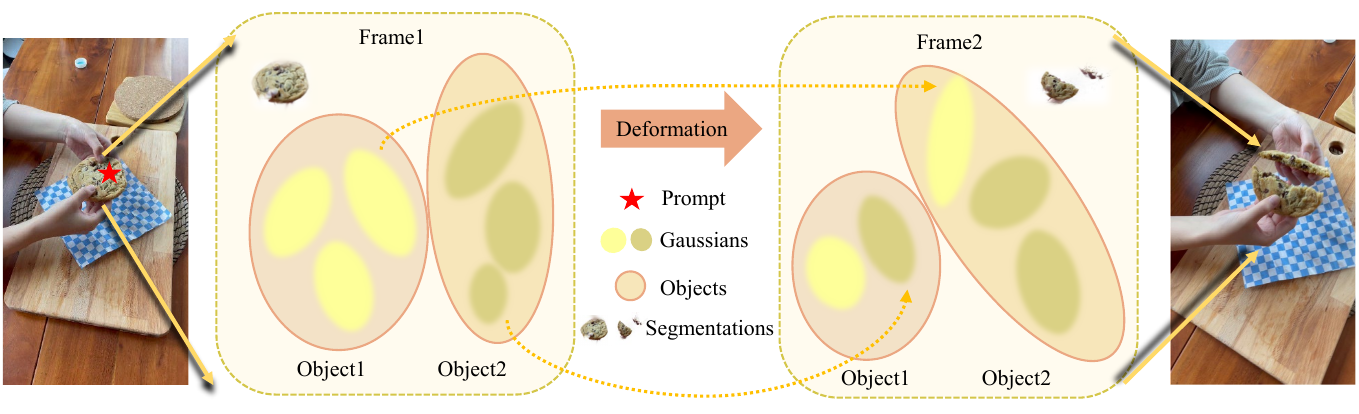}
    \caption{Illustration of \textit{Gaussian drifting} between objects in 4D-GS~\cite{wu20234daussians} on the HyperNeRF~\cite{park2021hypernerf} dataset. The left part is a random input view and the `star' stands for prompt. It is obvious that the segmentation results become inaccurate in different timestamps. It is because some 3D Gaussians of the cookie (object 1) segmented in frame 1 transform into another object (object 2) in frame 2 as shown in the right part.}

    \label{fig:motivation}

\end{figure}

The first problem in building a 4D segmentation framework is \textit{how to find an efficient solution to lift SAM to 4D representations?}  An intuitive solution to building a segmentation framework is to perform 3D segmentation~\cite{cen2023segmentany3dgs,hu2024segment,zhou2023feature} on the \textbf{deformed 3D Gaussians},  and then propagate it to the other timestamps using the deformation field of the 4D Gaussians. However, this approach may suffer from \textit{Gaussian drifting}: the non-rigid motion in which some 3D Gaussians belong to one object before timestamp $t_0$ but become part of another object afterward, as shown in Fig.~\ref{fig:motivation}.


To tackle the aforementioned challenges, we propose SA4D, which extends SAM to 4D representations. For precise 4D segmentation, we choose 4D Gaussian Splatting (4D-GS)~\cite{wu20234daussians} as the 4D representation. This method explicitly trains an independent Gaussian deformation field for motion and maintains a global canonical 3D Gaussians, allowing the use of sparse semantic features from any other frames at different timestamps, unlike other representations~\cite{4dgs-2, duan20244d} that focus only on local temporal spaces. Since the 4D world is typically captured by videos, we adopt a video object tracking foundation model \cite{DEVA} to generate per-frame identity masks. To efficiently compose sparse and noisy semantic priors, we propose a temporal identity feature field network, which (a) successfully resolves \textit{Gaussian drifting} by predicting time-dependent Gaussian semantic features, and (b) enables learning precise object identity information from noisy inputs of the video object tracking foundation model. After training, we introduce a Gaussian identity table to consolidate the segmentation results of 3D Gaussians at any training timestamps and conduct the post-processing on this table. For the temporal interpolation task, we adopt the nearest timestamp interpolation in the identity table. Our contribution can be summarized as follows: 
\begin{itemize}
    \item We reformulate the problem of 4D segmentation, and propose the Segment Any 4D Gaussians (SA4D) framework to lift the segment anything model to 4D representation with high efficiency.
    \item A temporal identity feature field includes a compact network that learns Gaussians' identity information across time from the noisy feature map input and eases the \textit{Gaussian drifting}. The segmentation refinement process also improves the inference rendering speed and makes scene manipulations simpler and more convenient. 
    \item SA4D achieves fast interactive segmentation within 10 seconds using an RTX 3090 GPU, photo-realistic rendering quality, and enables efficient dynamic scene editing operations seamlessly, e.g. removal, recoloring, and composition.
\end{itemize}


\section{Related Works}
\label{sec:related_work}
\textbf{3D/4D Representations.} Simulating real-world scenes has long been a subject of extensive research in the academic community. Many approaches~\cite{collet2015high,li20184d,guo2015robust,su2020robustfusion,flynn2019deepview,guo2019relightables,hu2022hvtr,li2017robust} are proposed to represent real-world scenes and achieve significant success. NeRF~\cite{mildenhall2021nerf,barron2021mip,zhang2020nerf++,wang2023neus2,michalkiewicz2019implicit,zhong2023color} and its extensions are proposed to render high-quality novel views even in sparse~\cite{liu2023zero,chen2023cascade}, multi-exposures~\cite{martin2021nerfinthewild,huang2022hdr,wu2024fast}, and show great potential in many downstream tasks~\cite{yuan2022nerfediting,cen2023segment,poole2022dreamfusion}. Grid-based representations~\cite{dvgo,fridovich2022plenoxels,xu20234k4d,gneuvox,tensor4d,gan2023v4d,msth} accelerate NeRF's training from days into hours even in minutes. Several methods also succeed in modeling dynamic scenes~\cite{tineuvox,hexplane,kplanes,tensor4d,lin2023im4d} but suffered on volume rendering~\cite{drebin1988volume}. Gaussian Splatting (GS)~\cite{3dgs,huang20242d,yu2023mip,yin20234dgen} based representations bring rendering speed into real-time while maintaining high training efficiency. Modeling dynamic scenes with Gaussian Splatting exists several ways: incremental translation~\cite{dynamic3dgs,sun20243dgstream}, temporal extension~\cite{duan20244d,4dgs-2,zhang2024togs} and global deformation~\cite{wu20234daussians,yang2023deformable3dgs, huang2023sc, li2023spacetime,gao2024gaussianflow,lin2023gaussian}. In this paper, we choose the global deformation GS representation, 4D-GS~\cite{wu20234daussians}, as our 4D representation because it owns a global canonical 3D Gaussians as its geometry and the ability to model monocular/multi-view dynamic scenes. The segmentation results on the deformed 3D Gaussians can also be transformed to the other timestamp easily.

\textbf{NeRF/Gaussian-based 3D Segmentation.} Prior to 3D Gaussians, numerous studies have extended NeRF to 3D scene understanding and segmentation. Semantic-NeRF~\cite{zhi2021place} first incorporated semantic information into NeRF and achieved 3D-consistent semantic segmentation from noisy 2D labels. Then, subsequent researches ~\cite{10044395, Kundu_2022_CVPR, 10203291, wang2023dmnerf} have developed object-aware implicit representations by introducing instance modeling but relying on GT labels. To achieve open-world scene understanding and segmentation, several approaches ~\cite{N3F,NEURIPS2022_93f25021,Kerr_2023_ICCV,Goel_2023_CVPR} distill 2D visual features from 2D foundation models~\cite{radford2021clip,li2022languagedriven,caron2021emerging} into radiance fields~\cite{zhou2023feature,qin2023langsplat}. However, these methods fail to segment semantically similar objects. Therefore, some approaches~\cite{cen2023segment,kim2024garfield,ye2023gaussiangrouping,cen2023segmentany3dgs} resorted to SAM's~\cite{kirillov2023segmentanything,ye2023gaussiangrouping} impressive open-world segmentation capability. The repair process~\cite{yang2024gaussianobject} can be followed to extract high-quality object representation after the segmentation process. Nevertheless, all the above methods are constrained to 3D static scenes. However, Directly utilizing the 3D segmentation methods on the 4D representations may also fall to \textit{Gaussian drifting}. Our methods solve the \textit{Gaussian drifting} by maintaining an identity encoding feature field, which models the deformation of semantic information.

\textbf{Dynamic Scene Segmentation.}
Few researchers have delved into dynamic scene segmentation. NeuPhysics~\cite{NEURIPS2022_53d3f457} only allows the complete segmentation of either the dynamic foreground or static background. Recently, 4D-Editor~\cite{jiang20234deditor} distills DINO features into a hybrid semantic radiance field and conducts 2D-3D feature match and Recursive Selection Refinement method per frame to achieve 4D editing. However, it takes 1-2 seconds to edit one frame. Moreover, ground truth masks of the dynamic foreground are needed to train the hybrid semantic radiance field. These limitations constrain its practical applicability. In this work, we propose a novel 4D segment anything framework enabling efficient dynamic scene editing operations. e.g. recoloring, removal, composition.


\begin{figure}[t]
    \centering
    \includegraphics[width=1\linewidth]{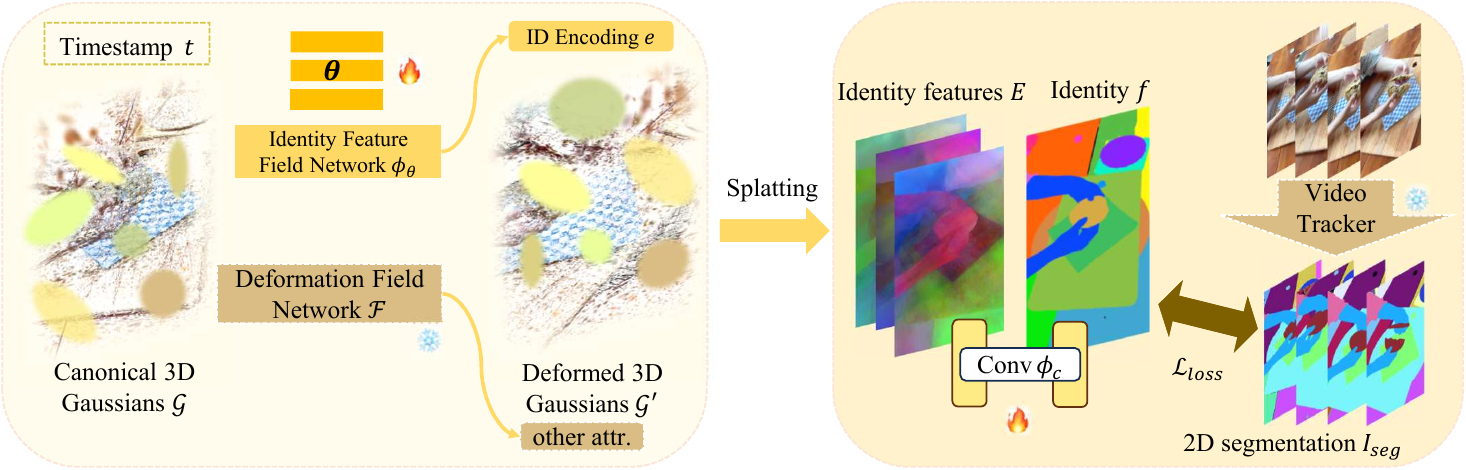}
    \caption{Overview of our training pipeline. Given a timestamp $t$ and canonical 3D Gaussians $\mathcal{G}$, the ID encoding $e$ and deformed 3D Gaussians $\mathcal{G}^{'}$ will be predicted by an optimizable $\phi_{\theta}$ and frozen deformation field network $\mathcal{F}$, respectively. Then the ID encoding $e$ is splatted to $E$, then $\phi_c$ is used to classify each pixel's ID $f$, and the whole training pipeline is supervised by $I_{seg}$ predicted by video tracker with $\mathcal{L}_{loss}$.}
    \label{fig:overall_pipeline}
    \vspace{-10pt}

\end{figure}


\section{Preliminaries}
\label{preminaries}

\subsection{Problem Formulation} 
Since there are few previous works that mainly focus on 4D segmentation, reformulation is necessary. It is worth noting that the 4D representation shows its weakness in segmentation, we define 4D segmentation as follows: 

\textbf{Problem:} Given any deformation-based 4D Gaussian representations $O$ trained on a dataset $\mathbf{L}$, The problem is 
to find an efficient solution $\mathcal{A}:(O,\mathbf{L}) \rightarrow O^{'}$, which $O^{'}$ is a set of object $O^{'} = \{o^{'}_i \mid i = 1, 2, \ldots, n\}$. The object $o_i$ should satisfy several features:

\textbf{Proposition 1: }Given any timestamp $t_i$, a deformed 3D-GS $\mathcal{G}^{'}_i \in G$ could be export by $o_i$. e.g:
\begin{equation}
    \forall o^{'}_i \in O^{'},\ export(o^{'}_i,t_i) \in G.
    \label{definition1}
\end{equation}
\textbf{Proposition 2: }No ambiguity between all the objects after export in a certain timestamp. e.g:
\begin{equation}
    \forall t_i, \{o^{'}_i, o^{'}_j\} \in  O^{'},\ export(o^{'}_i,t_i) \cap\ export(o^{'}_j,t_i) = \varnothing.
    \label{definition2}
\end{equation}
\textbf{Proposition 3: }Assume there exists a unique real-world 3D object \( o \), then for all \( t_i, o^{'}_i \in O^{'} \), we have
\begin{equation}
\forall t_i, o_i \in  O{'}, \quad export(o^{'}_i, t_i) = o.
\label{definition3}
\end{equation}
When rasterizing the object $o$ at any views $V=\{M,K\}$, the splatted image of $ \hat{I}_{seg}=\mathcal{S}(o,M,K)$ should correspond to ground truth ID segmentation $I_{seg}$:
\begin{equation}
    I_{seg} \sim \hat{I}_{seg} =  \mathcal{S}(o,M,K),
    \label{defination4}
\end{equation}
where $\mathcal{S}$ is differential splatting~\cite{yifan2019differentiablesplatting}, $M$ is extrinsic matrix and $K$ is camera intrisic matrix. Because of Eq.~\eqref{definition1} and Eq.~\eqref{definition3}, $o$ can be converted to $\hat{I}_{seg}$. Finally, we can select any object $o^{'}_i$ by the ID $\mathcal{P}$ from the $O{'}$.

\subsection{Gaussian Grouping}
Gaussian Grouping~\cite{ye2023gaussiangrouping} extends Gaussian Splatting to jointly reconstruct and segment anything in open-world 3D scenes. It introduces to each Gaussian a new parameter, identity encoding, to group and segment anything in 3D-GS~\cite{3dgs}. These identity encodings $e$ are then attached to the 3D Gaussians along with other attributes. Similar to rendering the RGB image in \cite{3dgs}, 3D Gaussians to a specific camera view and compute the 2D identity feature $E_{id}$ of a pixel $p$ by differential splatting algorithm~\cite{yifan2019differentiablesplatting} $\mathcal{S}$:
\begin{equation}
\label{eq:render_E}
    E_{p} = \mathcal{S}(\mathcal{G},M,K) = \sum_{i \in \mathcal{N}} e_{i} \alpha_{i} \prod_{j=1}^{i-1} (1-\alpha_{j}),
\end{equation}
where $e_{i}$, $\alpha_{i}$ represents the Identity Encoding and density of the $i$-th Gaussian under the view $\{M,K\}$. In \cite{ye2023gaussiangrouping}, a linear classifier $f$ segments the rendered image, which takes the rendered 2D identity feature map $E$ as inputs. Since 3D segmentation can be considered as a 4D scene with only one timestamp, \textbf{proposition 1,2,3} are also satisfied.

 \subsection{4D Gaussian Splatting} 4D Gaussian Splatting (4D-GS)~\cite{wu20234daussians} extends 3D-GS~\cite{3dgs} to model dynamic scenes efficiently, which represent 4D scene by a compact representation $O=\{\mathcal{G},\mathcal{F}_\theta(\mathcal{G},t)\}$. $\mathcal{G} \in G$ is a set of canonical 3D Gaussians belongs to 3D-GS $G$ and the Gaussian deformation field network $\mathcal{F}$ contains a spatial-temporal structure encoder $\mathcal{H}$ and a multi-head Gaussian deformation decoder $\mathcal{D}$. At timestamp $t$, the temporal and spatial features of 3D Gaussians are encoded through the spatial-temporal structure encoder:
\begin{equation}
    f_d=\mathcal{H}(\mathcal{G}, t).
\end{equation}
Then, the deformation decoder $\mathcal{D}=\left\{\phi_x, \phi_r, \phi_s \right\}$ employs three separate MLPs to compute the deformation of Gaussians' position, rotation, and scaling:
\begin{equation}
    (\Delta \mathcal{X}, \Delta r, \Delta s) = \mathcal{D}(f_d) = (\phi_x(f_d), \phi_r(f_d), \phi_s(f_d)).
\end{equation}
Finally, we can $export$ the deformed 3D Gaussians $\mathcal{G^{'}} = \{ \mathcal{X}^{'}, r^{'}, s^{'}, \sigma, \mathcal{C} \} = \{ \mathcal{X}+\Delta \mathcal{X}, r+\Delta r, s+\Delta s, \sigma, \mathcal{C} \}$ and render novel views $I = \mathcal{S}(\mathcal{G^{'}},M,K)$ via differential splatting $\mathcal{S}$~\cite{yifan2019differentiablesplatting} given any camera poses. To this end, we define the $export$ process of 4D-GS as:
\begin{equation}
    \mathcal{G}^{'}_i = export(O,t_i) = \mathcal{F}(\mathcal{G},t_i),
\end{equation}\label{export_4dgs}
which could satisfy the Eq.~\eqref{definition3}. It means that $\mathcal{G}$ are mapped to $\mathcal{G}_i^{'}$ conditioned on $t_i$ by $\mathcal{F}$. However, when conducting 4D segmentation, the vanilla 4D-GS pipeline doesn't contain any 4D semantic information.  Though canonical 3D Gaussians can be divided into a subset of 3D Gaussians $\mathcal{G}_{i}$ if adding $e_i$ to $g_i$ directly, which can satisfy the \textbf{proposition 1,2}, both canonical 3D Gaussians $\mathcal{G}_i$ and deformation field network $\mathcal{F}$ cannot identify each object and cause \textit{Gaussian drifting} as shown in Fig.~\ref{fig:motivation}. Therefore, the \textbf{proposition 3} is hard to be served. To address the problem, our SA4D framework is introduced.

\section{SA4D}
\label{sec:method}



\subsection{Overall Framework} Our key insight is introducing a representation to encode the temporal semantic information from a pretrained foundation model $\mathcal{V}$ to help the $export$ process since conduct segmentation in 4D-GS cannot afford the \textbf{proposition 3}. In SA4D, we refine the $export$ process as follows:
\begin{equation}
    \mathcal{G}^{'}_i = export(o_i,t_i) = \mathcal{F_\theta}(\mathcal{G}_i,t_i) \cap \mathcal{F}(\mathcal{G},t_i),
\label{eq:export}
\end{equation}
where $\mathcal{F}_\theta$ is our proposed SA4D representation including a temporal identity feature field $\phi_{\theta}$, tiny convolutional decoder $\phi_c$ and a Gaussian identity table $\mathcal{M}$ as shown in Eq.~\eqref{eq:framework_sa4d}:
\begin{equation}
    \mathcal{F}_\theta = f \cap M_i(f,I_{mask},t_i),
    \label{eq:framework_sa4d}
\end{equation}
in which $f$ is the Gaussian identity which is predicted by the identity encoding feature field as it is discussed in Sec~\ref{sec:identity_encoding_feature_field}, the latter is the post-process formula in Sec~\ref{sec:segmentation refinement}.

\subsection{Identity Encoding Feature Field}
\label{sec:identity_encoding_feature_field}

\textbf{Temporal Identity Feature Field Network.} To address the challenges of \textit{Gaussian drifting} discussed in Sec.~\ref{sec:intro}, we propose an identity feature field network $\mathcal{F_{\theta}}$ to encode identity features at each timestamp. Given time $t$ and center position $\mathcal{X}$ of 3D Gaussians $\mathcal{G}$ in the canonical space as inputs, the temporal identity feature field network predicts a low-dimensional time-variant Identity features $e$ for each Gaussian:
\begin{equation}
    e = \phi_{\theta}(\gamma(\mathcal{X}), \gamma(t))),
\end{equation}
where $\theta$ denotes the learnable parameters of the network and $\gamma$ denotes the positional encoding~\cite{vaswani2017attention}. Then, identity splatting as shown in Eq.~\eqref{eq:render_E} is rendered to get $E_p$. Note that the supervision is the identity of each pixel, we use a tiny convolutional decoder $\phi_c$ and the softmax function to predict Gaussian identity $f$ as shown in Eq.~\eqref{eq:predict_identity}:
\begin{equation}
    f = softmax(\phi_c(E_p)).
    \label{eq:predict_identity}
\end{equation}
\textbf{Optimization.} Since it is hard to access GT 4D object labels, we cannot supervise the training process with $o$. Therefore, we adopt the 2D pseudo segmentation results of foundation video tracker $I_{mask}$ as the supervision. Similar to ~\cite{ye2023gaussiangrouping}, we use the 2D Identity Loss $L_{2d}$ and 3D Regularization Loss $L_{3d}$ to supervise the training process $L = \lambda_{2d}L_{2d} + \lambda_{3d}L_{3d}$. Identity loss $L_{2d}$ is a standard cross-entropy loss on the training image $I$:
\begin{equation}
    L_{2d} = -\frac{1}{||I||} \sum_{i \in I} \sum_{c=1}^{C} {p(i) \log \hat{p}(i)},
    \label{eq:CELoss}
\end{equation}
where $C$ is the total number of mask identities in $I_{mask}$ and $p$ and $\hat{p}$ are the multi-identity probability of gt mask $I_{mask}$ and our network predictions.
3D Regularization Loss $L_{3d}$ in \cite{ye2023gaussiangrouping} is applied to further supervise the Gaussians inside 3D objects and heavily occluded. It is the KL divergence loss, enforcing the Identity Encodings of the top $k$-nearest 3D Gaussians to be close in their feature distance:
\begin{equation}
    L_{3d} = \frac{1}{m} \sum_{j=1}^{m}{D_{KL}(P||Q)} = \frac{1}{mk}\sum_{j=1}^{m}{\sum_{i=1}^{k}{f_{j} \log \frac{f_{j}}{f_{i}}}},
    \label{eq:unsupervise3d}
\end{equation}
where $m$ is the number of sampled Gaussians, $P$ is the $j$-th Gaussian and $Q$ is the set of its $k$-nearest neighbors in the deformed space.

\subsection{4D Segmentation Refinement}
\label{sec:segmentation refinement}
\textbf{Post Processing.} Though the temporal identity feature field network shows the ability to encode temporal identity features of 3D Gaussians, the $export$ process is still affected by the heavily occluded and invisible Gaussians and the noisy segmentation supervision. To get a more precise $\mathcal{G}^{'}_i$ which is similar to $o$, we employ 2-step post-processing. The first step is removing outliers, the same as ~\cite{cen2023segmentany3dgs,ye2023gaussiangrouping}. However, there are still some ambiguous Gaussians at the interface between two objects, as discussed in \cite{cen2023segment}, which do not or slightly affect the quantitative results but impair the geometry correctness of the segmentation target $o^{'}$. Therefore, in the second step, similar to \cite{cen2023segment}, we utilize the 2D segmentation supervision $I_{mask}$ as prior to eliminate these ambiguous Gaussians. Concretely, we assign each Gaussian $g$ a mask $m=\mathbb{I}(g \in o^{'})$, render 3D point masks and apply the mask projection loss in \cite{cen2023segment} :
\begin{equation}
    L_{proj} = - \sum_{i \in I}{M_{o^{'}}(i)} \cdot M(i) + \lambda \sum_{i \in I}{(1 - M_{o^{'}}(i)) \cdot M(i)},
    \label{eq: mask projection loss}
\end{equation}
where $M_{o^{'}}$ denotes the mask of $o^{'}$ in $I_{mask}$, $M$ is the rendered object mask and $\lambda$ is a hyper-parameter to determine the magnitude of the negative term. The Gaussians with negative gradients are then removed. The greater $\lambda$ is, the more sensitive $L_{proj}$ is to the ambiguous Gaussians.

\textbf{Gaussian Identity Table.} Going through the implicit identity encoding field network and performing post-process frame-by-frame during inference are time-consuming and inconvenient for scene editing. Therefore, we propose to store the segmentation results at each training timestamp in a Gaussian identity table $\mathcal{M}$ and employ the nearest timestamp interpolation $nearest$ during inference as shown in Eq.~\eqref{id_table}:
\begin{equation}
    M_i = nearest(\mathcal{M}(g,t_i)).
    \label{id_table}
\end{equation}
Concretely, after training, users can input an object ID. Our method segments out 4D Gaussians that belong to the object at each training timestamp according to their Identity Encoding.  The final segmentation results are stored in the Gaussian identity table. The post-process procedure can be applied in the $\mathcal{M}$ before rendering. This process can be finished within 10 seconds in most cases and increase the rendering speed during inference significantly. The details of the 4D segmentation refinement are in Algorithm.~\ref{algorithm:refine}.


\section{Experiment}
\label{sec:experiment}
\begin{table}[]
    \centering
    \begin{tabular}{c cc cc}
        \toprule
         \multirow{2}{*}{Method} & \multicolumn{2}{c}{HyperNeRF~\cite{park2021hypernerf}} & \multicolumn{2}{c}{Neu3D~\cite{li2022neural}}  \cr  
        \cmidrule(lr){2-3}\cmidrule(lr){4-5}
         ~ & mIoU(\%) & mAcc(\%) & mIoU(\%) & mAcc(\%) \\
         \midrule
         SAGA~\cite{cen2023segmentany3dgs} & 65.25 & 75.56 & 76.26 & 81.56 \\
         Gaussian Grouping~\cite{ye2023gaussiangrouping} & 69.53 & 91.55 & 87.02 & 98.72 \\
         \midrule
         Ours w/o TFF (w/o Refinement)& 80.26 & \textbf{99.56} & - & -  \\ 
         Ours w/ TFF (w/o Refinement) & 81.10 & 99.54 & 80.14 & \textbf{99.88}  \\
         Ours w/ all& \textbf{89.86} & 99.24 & \textbf{93.02} & 99.76 \\
         \bottomrule
    \end{tabular}
    \vspace{5pt}
    \caption{Evaluation metrics on the HyperNeRF~\cite{park2021hypernerf} and Neu3D~\cite{li2022neural} dataset. ``Ours w/o TFF'' means directly  attaching an identity feature vector to each Gaussian.}
    \label{tab:average}
\end{table}

\subsection{Experimental setups}\label{experimental setups}
\textbf{Implementation Details.} Our implementation is primarily based on the PyTorch~\cite{NEURIPS2019_bdbca288} framework and tested in a single RTX 3090 GPU. For simplicity, we train our framework based on a pre-trained 4D-GS~\cite{wu20234daussians}. The output dimension of the temporal identity feature field network, e.g. Identity Encoding dimension, is set to 32. The classification convolutional layer has 32 input channels and 256 output channels. We use the Adam optimizer for both the temporal identity feature field network and convolutional layer, with $5000$ training iterations and a learning rate of $5e-4$. The training process costs about half an hour for a standard scene with 200k Gaussians under the resolution of 1352$\times$1014. For optimization loss, we choose $\lambda_{2d}=1$, $\lambda_{3d}=2$, $k=5$, and $m=1000$, the same as~\cite{ye2023gaussiangrouping}. Note that we only use images under one camera as training inputs for Neu3D~\cite{li2021neural} scenes because of the ID conflict problem produced by \cite{DEVA}. We leave the problem as our future work.

\textbf{Datasets.} We use two widely-used dynamic scene datasets in our experiments, which are HyperNeRF dataset~\cite{park2021hypernerf} and Neu3D dataset~\cite{li2022neural}. As there are no ground truth segmentation mask labels for these datasets, we annotate some dynamic objects in test views in these scenes for our evaluation. Details about mask annotation are provided in the Supp. file.


\textbf{Baselines.} Due to the lack of open source code for \cite{jiang20234deditor} and no other 4D segmentation methods, we choose 3D segmentation methods as our baselines. Most recent Gaussian-based 3D segmentation methods also lack open source code and Feature-3DGS~\cite{zhou2023feature} mainly focuses on semantic segmentation and language-driven editing, so we choose SAGA~\cite{cen2023segmentany3dgs} and Gaussian Grouping~\cite{ye2023gaussiangrouping} for comparison. Since they do not have the temporal dimension and are limited to 3D static scenes, we conduct training and segmentation on a certain training timestamp based on the \textbf{deformed 3D Gaussians} in 4D-GS~\cite{wu20234daussians}. The segmentation results are then propagated to other timestamps and viewed through 4D-GS's deformation field. We train the baseline models on 10 randomly selected timestamps separately for each scene and reported the average test results.

\begin{figure}[t]
    \centering
    \includegraphics[width=1\linewidth]{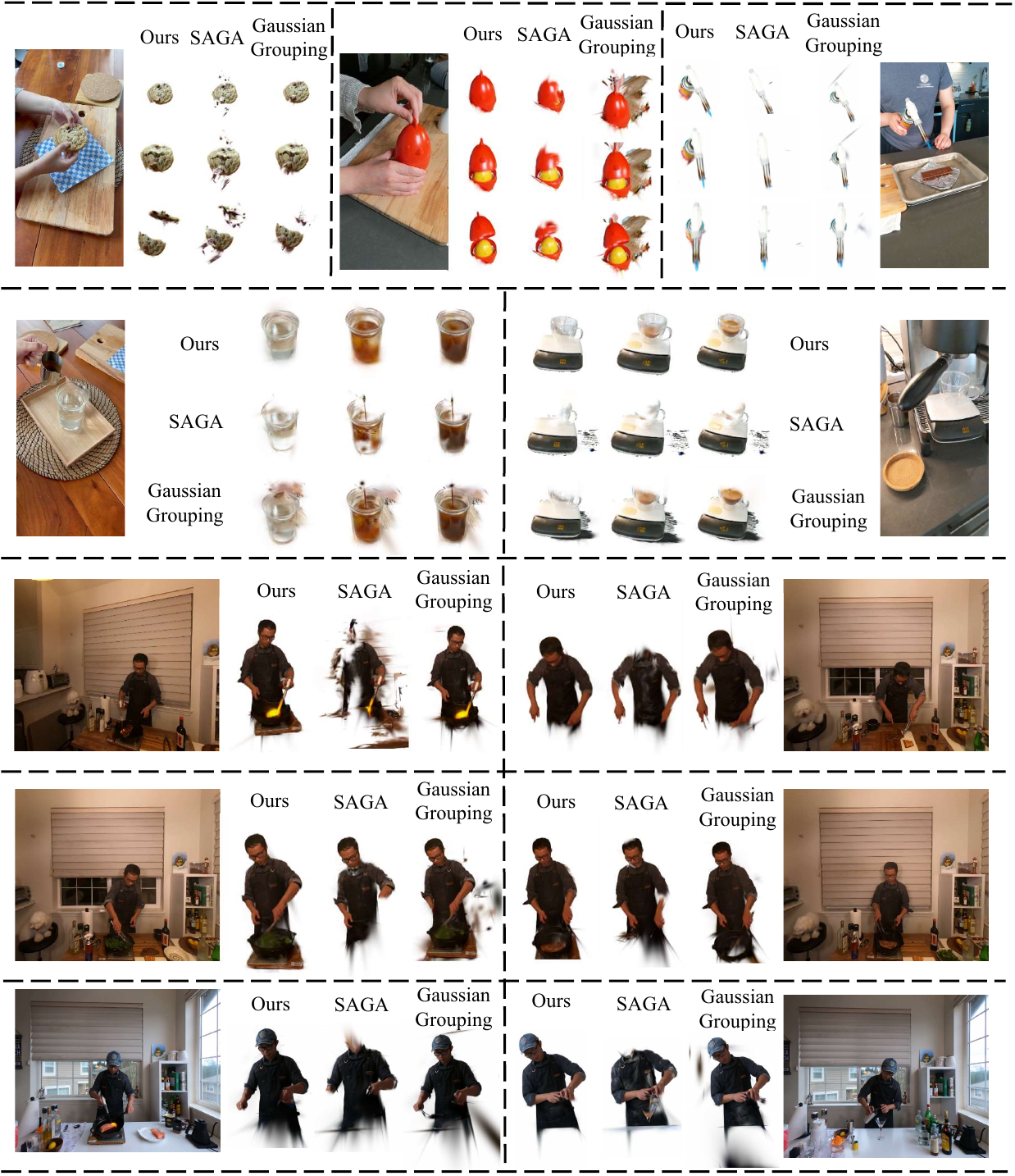}
    \vspace{-15pt}
    \caption{Visual comparisons of our method and baselines on the HyperNeRF~\cite{park2021hypernerf} and Neu3D~\cite{li2022neural} dataset. The upper five scenes are from the HyperNeRF dataset and we visualize and compare the segmentation results at three random novel views and timestamps for each scene. The lower six scenes are from the Neu3D dataset and we visualize and compare the segmentation results at one random novel view and timestamp for each scene.}
    \label{fig:Results}
\end{figure}

\subsection{Results}
\label{subsec: results}


Since acquiring and annotating 3D/4D Gaussian labels is significantly challenging, we render 3D point masks to 2D, threshold the rendered 2D mask value to be greater than 0.1 to remove the low-density areas that contribute negligibly to the rendered visuals and calculate the IoU and Accuracy. The averaged quantitative results are provided in Tab.~\ref{tab:average} and per-scene results are provided in Tab.~\ref{tab:hypernerf results} and Tab.~\ref{tab:dynerf results}. Visual comparisons are shown in Fig.~\ref{fig:Results} and more visualization results are provided in Fig.~\ref{fig: more visualizations 1}, Fig.~\ref{fig: more visualizations 2} and Fig.~\ref{fig: more visualizations 3}.
\begin{wraptable}{r}{0.4\linewidth}
    \centering
    \resizebox{1.0\linewidth}{!}{
    \begin{tabular}{c | c c c}
    \toprule
    Model & FPS & Storage(MB) \\
    \midrule
    Ours w/o. GIT & 7.12 & 62.55 \\
    Ours & 40.36 & 80.12 \\
    \bottomrule
    \end{tabular}
    }
    \vspace{-10pt}
    \caption{Ablation of the Gaussian Identity Table.}
    \label{tab:GIT}
\end{wraptable}
The quantitative results on HyperNeRF~\cite{park2021hypernerf} scenes are shown in Tab.~\ref{tab:hypernerf results}. In the monocular setting, existing 3D segmentation methods can only rely on one view at one timestamp rather than multi-view observations. Therefore, the two baselines produce terrible segmentation results.  The reason that the Gaussian Grouping baseline has low mIoU but high mACC is that it incorporates too much noise due to the heavy occlusion and monocular sparsity during training. The baselines also suffer from the \textit{Gaussian Drifting} problem as shown in the \textit{split-cookie} scene in Fig.~\ref{fig:Results}. Although 3D segmentation models~\cite{cen2023segmentany3dgs}~\cite{ye2023gaussiangrouping} successfully segment out the 3D Gaussians of the cookie at the first frame and track their motions, one piece of the cookie disappears later. In contrast, our method can accurately identify the 3D Gaussians belonging to the cookie at different timestamps remedying the \textit{Gaussian Drifting} problem.

The quantitative results on Neu3D scenes are shown in Tab.~\ref{tab:dynerf results}. The Neu3D datasets are less challenging than the monocular datasets because 4D-GS~\cite{wu20234daussians} can model the motion more accurately in the multi-view setting and 3D segmentation models have multi-view observations at one timestamp. Also, the \textit{Gaussian Drifting} problem is not obvious. However, our method still has superior performance over the two baselines by learning temporal identity features. SAGA~\cite{cen2023segmentany3dgs}'s performance is not stable because it is class-agnostic and it struggles to segment large objects with mutiple semantics. Utilizing multi-view information, Gaussian Grouping~\cite{ye2023gaussiangrouping} sometimes outperforms ours w/o refinement. Note that since we only train our model with temporal mask identities in a single view, the initial segmentation results before refinement are sometimes noisy.    

\begin{figure}[t]
    \centering
    \includegraphics[width=1\linewidth]{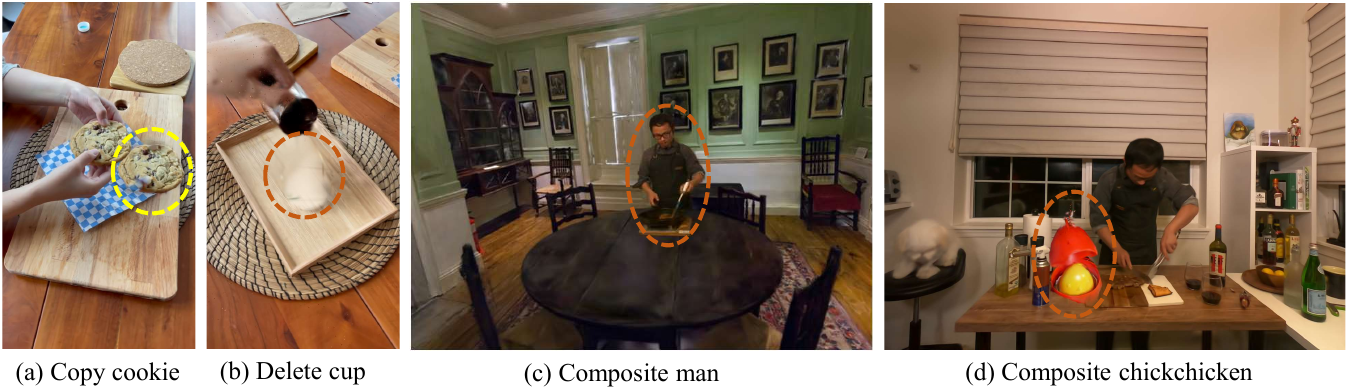}
    \vspace{-10pt}
    \caption{More examples of composition, deletion with segmented 4D Gaussians. (a): copying a cookie in the scene. (b) Deleting the cup in the scene. (c) Compositing the man with a room in Neu3D~\cite{li2021neural} and Mip-NeRF360~\cite{mipnerf} dataset. (d) Compositing the chickchicken with the man in HyperNeRF~\cite{park2021hypernerf} and Neu3D~\cite{li2021neural} dataset.}
    \label{fig:scene edit}
\end{figure}


\begin{figure}[t]
    \centering
    \includegraphics[width=1\linewidth]{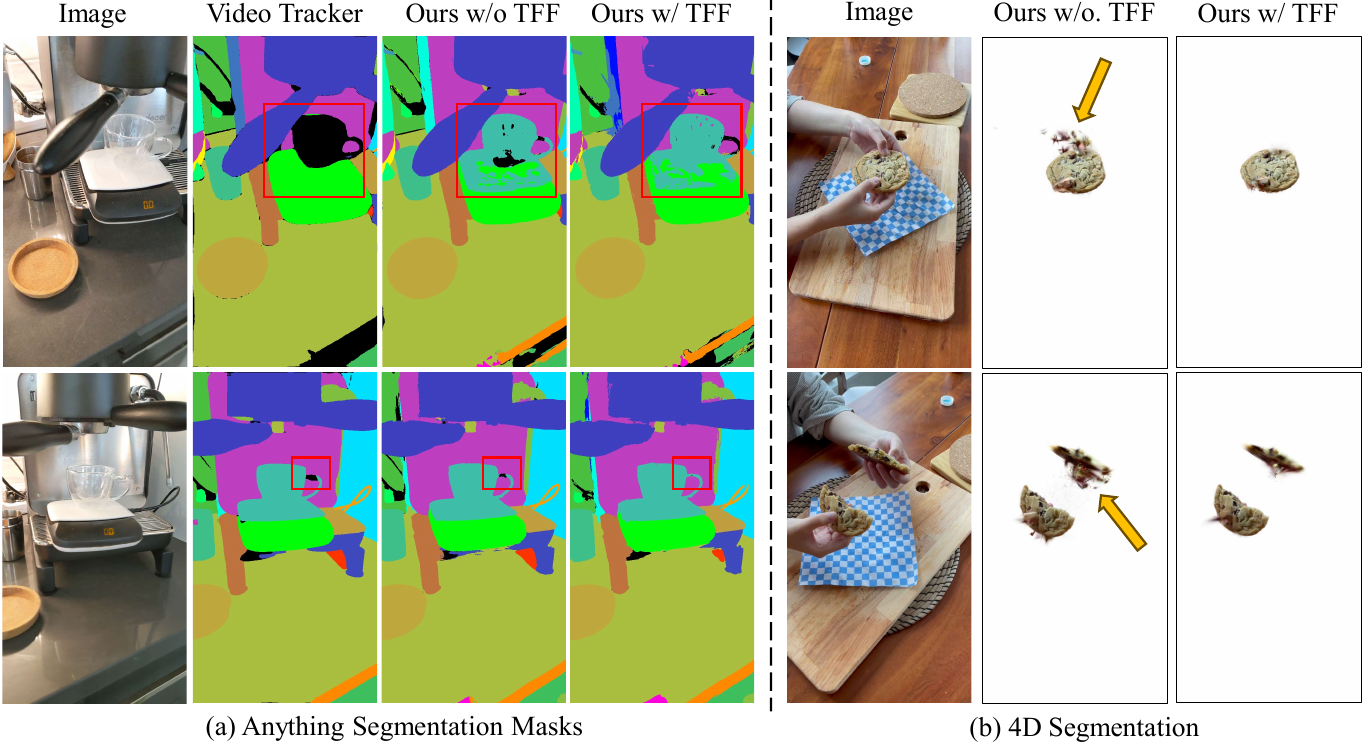}
    \vspace{-20pt}
    \caption{Ablation study of our temporal identity field network. (a) The black regions represent the void class (Illustrated as black color). Predictions from input 2D supervision (e.g. video tracker~\cite{DEVA}) are sometimes incorrect (e.g. cup labeled void 0 in the image above) and noisy (e.g. handle labeled void in the image below). (b) Due to the \textit{Gaussian drifting}, some Gaussians outside the cookie in the image above will transform into the cookie as shown in the image below.}
    \label{fig:ablation tff}
\end{figure}

\subsection{Dynamic Scene Editing}
In Fig.~\ref{fig:teaserfig} and Fig.~\ref{fig:scene edit}, we show some applications of our SA4D framework.
Thanks to the explicity of the 4D Gaussian representation and our Gaussian identity table, we can retrieve the 3D Gaussians of an object at each timestamp in real time and then manipulate them easily and quickly. For example, the object can be copied in the 4D scene as shown in column (a) of Fig.~\ref{fig:scene edit}, removed as shown in column (b), composited with 3D Gaussians as shown in column (c) or composited with other 4D Gaussians in column (d).

\subsection{Ablation Study}

\begin{figure}[t]
    \centering
    \includegraphics[width=1\linewidth]{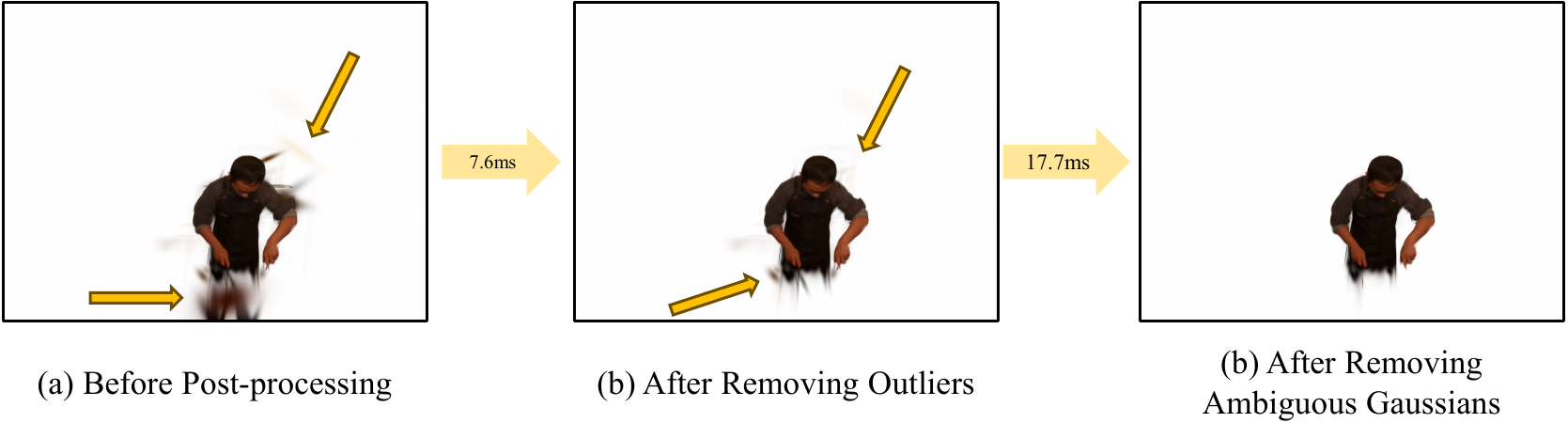}
    \vspace{-20pt}
    \caption{Ablation study on the effect of the post-processing in 4D segmentation refinement. Time in the arrows stands for the average time consumption in each frame. }
    \label{fig: ablation post-process}
\end{figure}

\textbf{Temporal Identity Feature Field Network.} 
Instead of attaching a time-invariant feature vector to each Gaussian as most 3D segmentation methods do, we introduce a Temporal Identity Feature Field Network (TFF) to encode identity features across time to solve the Gaussian drifting. We compare the two methods on monocular datasets, HyperNeRF~\cite{park2021hypernerf}, where it is challenging for 4D-GS~\cite{wu20234daussians} to model motion accurately. Note that we did not conduct post-processing in our ablation study of TFF in order to study the effect of Gaussian drifting. As shown in Tab.~\ref{tab:hypernerf results}, ours w/ TFF outperforms ours w/o TFF on most scenes, especially on the split-cookie scene where we observe severe Gaussian drifting. We also found that our temporal identity field shows the ability to learn accurate ID information from noisy input 2D segmentation supervision. Visual comparisons and illustrations are shown in Fig.~\ref{fig:ablation tff}.

\begin{wraptable}{r}{0.4\linewidth}
    \centering
    \resizebox{1.0\linewidth}{!}{
    \begin{tabular}{c | c c c}
    \toprule
    Interval & Time(s) & IoU(\%) & Acc(\%) \\
    \midrule
    1 & 7.8 & 88.92 & 99.73 \\
    2 & 3.91 & 88.91 & 99.74\\
    4 & 1.82 & 88.61 & 99.66 \\
    8 & 0.90 & 88.11 & 99.51 \\
    16 & 0.48 & 87.58 & 99.07\\
    \bottomrule
    \end{tabular}
    }
    \vspace{-10pt}
    \caption{Ablation on refinement interval. “Time” in the table represents the time cost by 4D segmentation refinement.}
    \label{tab:ablation_interval}
\end{wraptable}

\textbf{4D Segmentation Refinement.} Since the segmentation supervision is in 2D, the initial segmentation result is typically noisy, including artifacts far away from the object and ambiguous Gaussians at the boundary fitting the color of multiple objects. As shown in Tab.~\ref{tab:average} and Fig.~\ref{fig: ablation post-process}, our refinement approach can obviously improve the segmentation quality and eliminate artifacts and ambiguous Gaussians at the boundary. Note that ambiguous Gaussians at the boundary partially overlap with the object, so removing them may result in a slight decrease in segmentation accuracy. We conduct ablation studies on the `split-cookie' in the HyperNeRF~\cite{park2021hypernerf} dataset. Tab.~\ref{tab:GIT} shows that the Gaussian Identity Table we proposed can maintain the real-time rendering speed of 4D-GS while not increasing too much storage. We also study the effect of the timestamp interval in 4D segmentation refinement. Tab.~\ref{tab:ablation_interval} shows that with the interval increasing, although the refinement time decreases, the segmentation quality downgrades. 



\section{Limitation}
\label{limitations}

Though SA4D can achieve fast and high-quality segmentation in 4D Gaussians, some limitations exist and can be explored in future works. (1) Similar to Gaussian Grouping~\cite{ye2023gaussiangrouping}, selecting an object needs an identity number as prompt, which causes difficulty for selecting desired object comparing with `click' or language. (2) The deformation field cannot be decomposed at the object level, necessitating the involvement of the entire deformation network in the segmentation and rendering process. (3) The mask identity conflicts between different video inputs make it difficult to utilize multi-view information effectively. (4) Similar to 3D segmentation, object artifacts still exist due to the features of 3D Gaussians.

\section{Conclusion} 
\label{conclusions}
This paper proposes a Segment Any 4D Gaussians framework to achieve fast and precise segmentation in 4D-GS~\cite{wu20234daussians}. The semantic supervision from world space in different timestamps is converted to canonical space by the 4D-GS and temporal identity feature field network. The temporal identity feature field network also solves the \textit{Gaussian drifting} problem. SA4D can render high-quality novel view segmentation results and also supports some editing tasks, such as object removal, composition, and recoloring.
{\small
\bibliographystyle{unsrt}

}

\clearpage


\appendix

\section{Appendix / supplemental material}
\subsection{Introductions}
The appendix provides some implementation details and further results that accompany the paper.
\begin{itemize}
    \item Section ~\ref{tff architecture} introduces the implementation details of our temporal identity field network architecture.
    \item Section ~\ref{segmentation process} provides more descriptions including pseudocode of our approach.
    \item Section ~\ref{more results} provides per-scene quantitative results.
    \item Section ~\ref{more visualizations} provides more visualization results.
    \item Section ~\ref{more limitations} discusses the limitations of our approach.
    \item Section ~\ref{mask annotation} provides the implementation details and visualizations of mask annotation.
    \item Section ~\ref{more discussions} provides more discussions.
\end{itemize}

\subsection{Network Architecture of the Temporal Identity Field}
\label{tff architecture}

\begin{wrapfigure}{r}{0.5\linewidth}
    \centering
    \includegraphics[width=\linewidth]{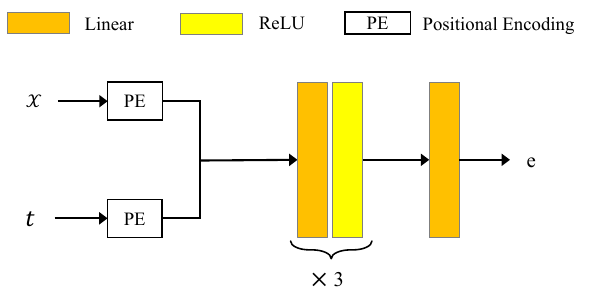}
    \vspace{-20pt}
    \caption{The architecture of our temporal identity field network.}
    \label{fig: tff architecture}
\end{wrapfigure}


The architecture of the temporal identity field network $\mathcal{F}_{\theta}$ are shown in Fig.\ref{fig: tff architecture}. The network is a simple MLP that takes time and the spatial positions of 3D Gaussians as inputs and outputs their corresponding identity encoding: $\mathcal{F}_{\theta}(\gamma(\mathcal{X}), \gamma(t)) = e$, where $\gamma$ denotes the positional encoding. Specifically, the network concatenates $\gamma(\mathcal{X})$ and $\gamma(t)$, passes them through three fully connected ReLU layers, each with 256 channels, and outputs a 256-dimensional feature vector. Then, an additional fully connected layer is applied to map the feature vector to the final 32-dimensional identity encoding.

\subsection{More Algorithm Descriptions}
\label{segmentation process}
We outline the pseudocode of training and 4D segmentation refinement for our SA4D in Algorithm~\ref{algorithm:training} and Algorithm~\ref{algorithm:refine}. 

\subsection{More Results}
\label{more results}
In Tab.~\ref{tab:hypernerf results} and Tab.~\ref{tab:dynerf results}, we provide the results for individual scenes associated with Sec.~\ref{subsec: results} of the main paper.

\subsection{More Visualizations}
\label{more visualizations}
\begin{figure}[t]
    \centering
    \includegraphics[width=1\linewidth]{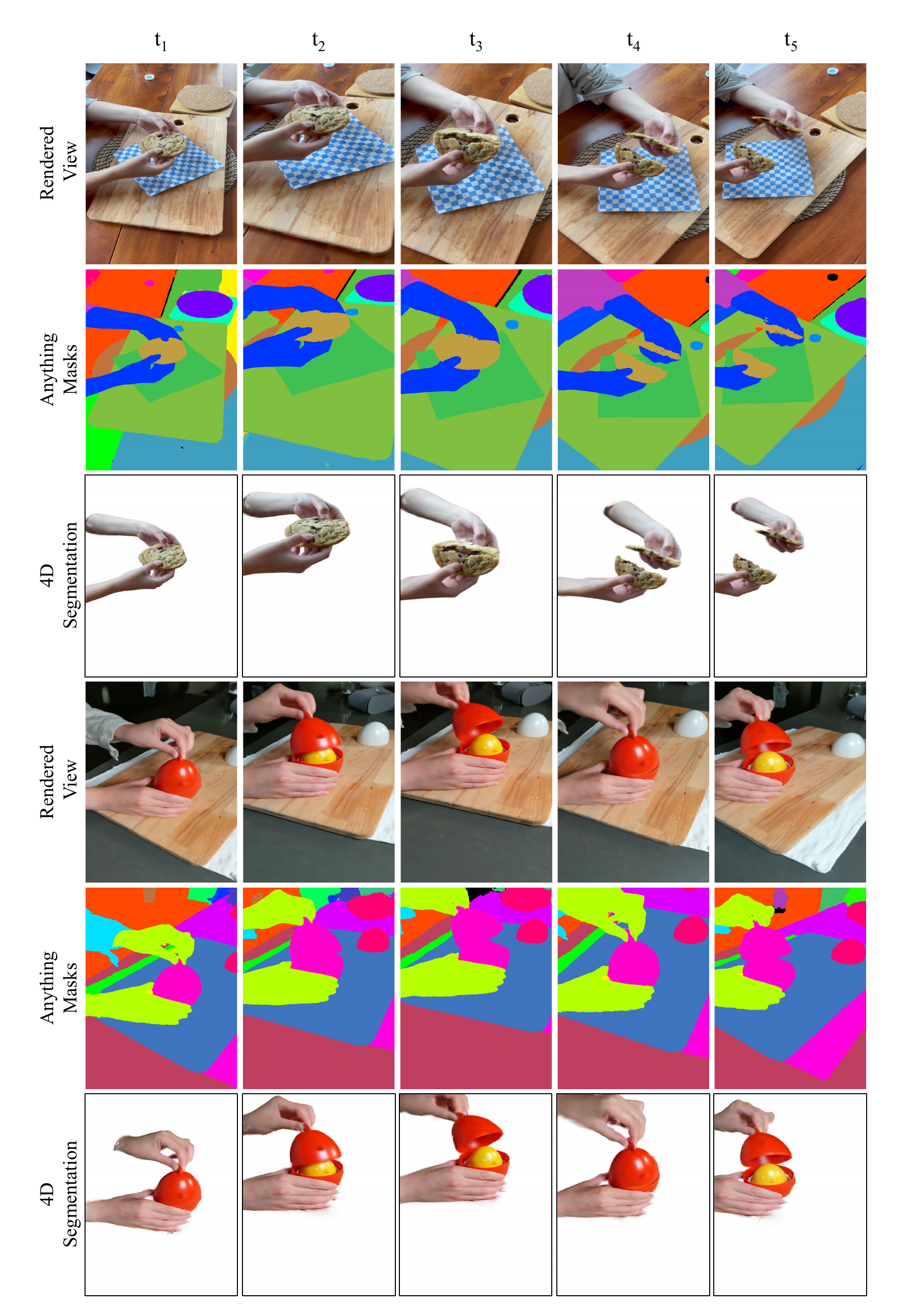}
    \vspace{-15pt}
    \caption{More visualizations results.}
    \label{fig: more visualizations 1}
\end{figure}

\begin{figure}[t]
    \centering
    \includegraphics[width=1\linewidth]{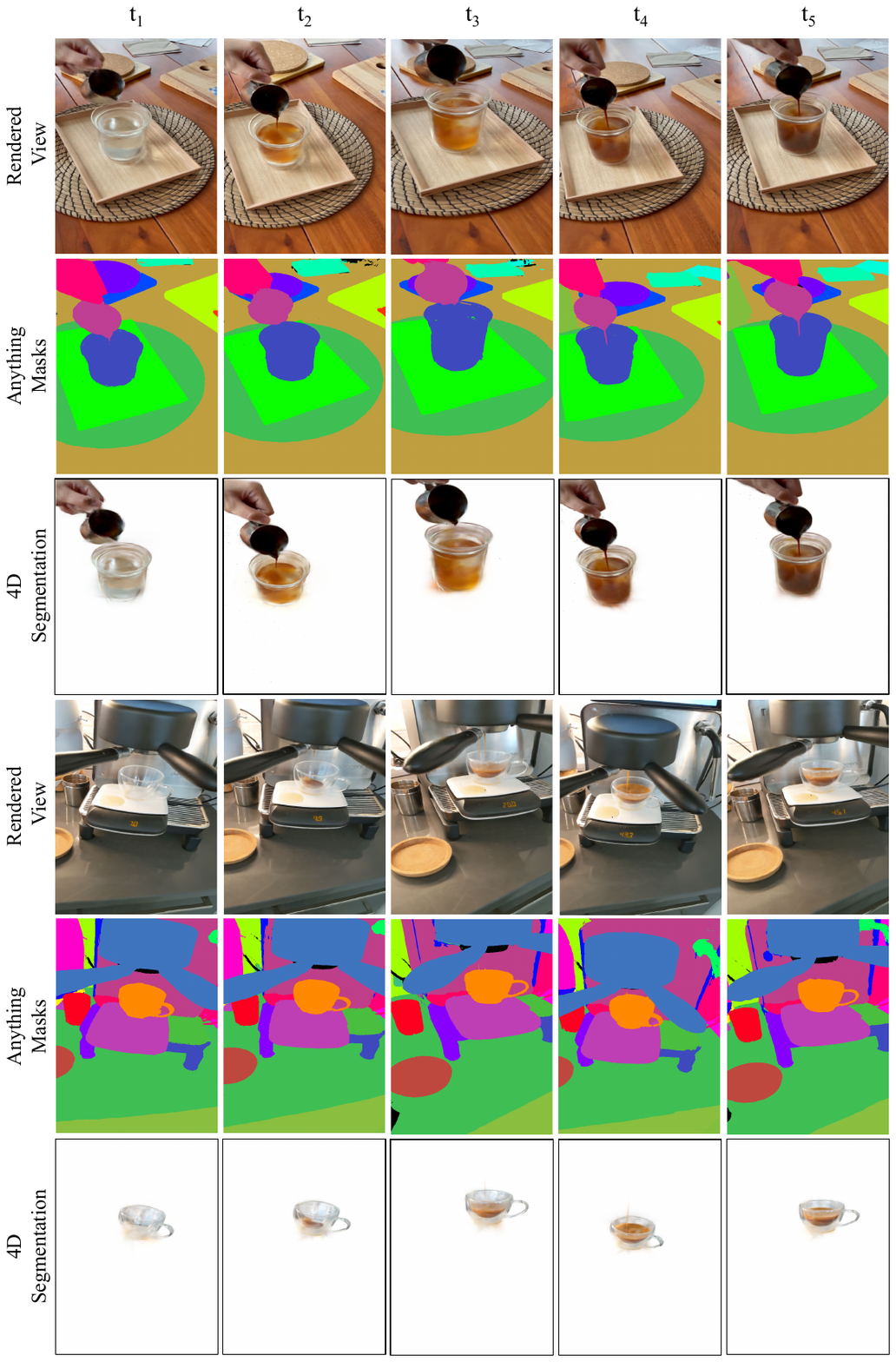}
    \vspace{-15pt}
    \caption{More visualizations results.}
    \label{fig: more visualizations 2}
\end{figure}

\begin{figure}[t]
    \centering
    \includegraphics[width=1\linewidth]{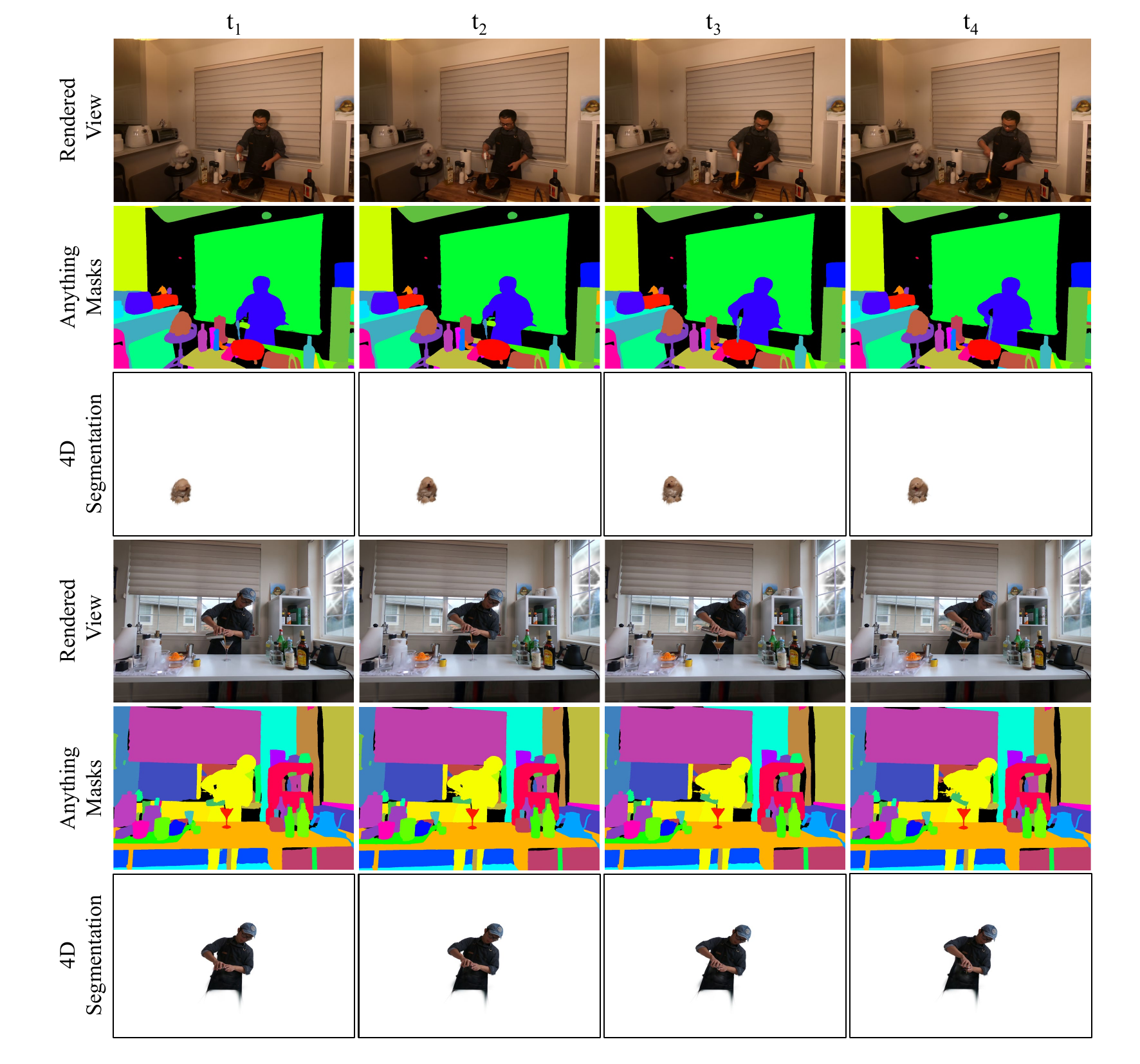}
    \vspace{-15pt}
    \caption{More visualizations results.}
    \label{fig: more visualizations 3}
\end{figure}

We present more visualization results in Fig.~\ref{fig: more visualizations 1}, Fig.~\ref{fig: more visualizations 2} and Fig.~\ref{fig: more visualizations 3}, showing the effectiveness of our
method. Our method can not only render anything masks at novel views and novel timestamps but also segment single or multiple objects in 4D dynamic scenes.

\begin{algorithm}
    \caption{SA4D Training Framework}
    \label{algorithm:training}
    \begin{algorithmic}
        \State $(\hat{M_1}, \hat{M_2},..., \hat{M_N}) \leftarrow$ Zero-shot Video Segmentation \Comment{Pseudo Label}
        \State $\mathcal{G}, \mathcal{F} \leftarrow$ InitPretrained4DGS() \Comment{Canonical 3D Gaussians and Deformation Field Network}
        \State $\phi_{\theta}, \phi_c \leftarrow$ Init() \Comment{Temporal Identity Field Network and Convolutional Decoder}
        \State $i \leftarrow 0$ \Comment{Iteration Count}
        \While{not converged}
            \State $V, t, \hat{I}, \hat{M} \leftarrow$ SampleTrainingView() \Comment{Camera View, Timestamp, Image and Mask}
            \State $\mathcal{G}^{'} \leftarrow \mathcal{F}(\mathcal{G}, t)$ \Comment{Deformed Gaussians}
            \State $e \leftarrow \phi_{\theta}(\gamma(\mathcal{X}), \gamma(t))$ \Comment{Compute Identity Encoding}
            \State $E \leftarrow$ Rasterize($\mathcal{G}^{'}$, $e$) \Comment{Rendered Identity Encoding}
            \State $L \leftarrow \lambda_{2d}L_{2d}(E, \hat{M}) + \lambda_{3d}L_{3d}(e)$ \Comment{Optimization Loss, see Eq.~\eqref{eq:CELoss} and Eq.~\eqref{eq:unsupervise3d} of the paper}
            \State $\phi_{\theta}, \phi_c \leftarrow$ Adam($\nabla L$) \Comment{Backprop \& Step}
            \State $i \leftarrow i + 1$.
        \EndWhile
        \State \Return $\mathcal{G}$
    \end{algorithmic}
\end{algorithm}

\begin{algorithm}
    \caption{4D Segmentation Refinement}
    \label{algorithm:refine}
    \begin{algorithmic}
        \State \textbf{Input:} Temporal identity field network $\phi_\theta$, Classifier head $\phi_c$, canonical 3D Gaussians $\mathcal{G}$, deformation field network $\mathcal{F}$, selected object IDs $l$, timestamp interval $k$, training images $\hat{I}$, corresponding camera views $V$, timestamps $t$ and segmentation masks $\hat{M}$
        \State \textbf{Output:} Gaussian identity table $\mathcal{M}$.
        \Procedure{Refinement}{$\phi_\theta, \phi_c, \mathcal{G}, \mathcal{F}, l, k, \hat{I}, V, t, \hat{M}$}
        \State $i \leftarrow 0$ \Comment{Iteration Count}
        \While{$i \leq N$} \Comment{Enumerate training views}
        \State $V_i, t_i, \hat{I}_i, \hat{M}_i \leftarrow$ SampleTrainingView(i) \Comment{Camera View, Timestamp, Image and Mask}
        \State $\mathcal{G}^{'} \leftarrow \mathcal{F}(\mathcal{G}, t_i)$ \Comment{Deformed Gaussians}
        \State $e \leftarrow \phi_{\theta}(\gamma(\mathcal{X}), \gamma(t_i))$ \Comment{Compute Identity Encoding}
        \State $\hat{\mathcal{G}} \leftarrow \phi_c(e)$ \Comment{Classified 3D Gaussians}
        \State $\mathcal{G}_l \leftarrow$ SelectGaussians($\hat{\mathcal{G}}$, $l$) \Comment{Segment target Gaussians}\
        \State $\mathcal{G}_l \leftarrow$ RemoveOutliers($\mathcal{X}$, $\mathcal{G}_l$) \Comment{Remove Outliers}
        \State $m \leftarrow \mathbb{I}(g \in \mathcal{G}_l)$  \Comment{Initialize 3D Point Masks}
        \State $M \leftarrow$ Rasterize($\mathcal{G}^{'}$, m) \Comment{Rendered Object Mask}
        \State $L \leftarrow L_{proj}(M, \hat{M}_i)$ \Comment{Mask Projection Loss, see Eq.~\eqref{eq: mask projection loss} of the paper}
        \State $G_m \leftarrow \nabla L$ \Comment{Gradients of 3D Point Masks}
        \For{$g_m$ in $G_m$} \Comment{Remove Ambiguous Gaussians at the Boundary}
        \If{$g_m < 0$}
        \State $\mathcal{G}_l \leftarrow$ RemoveGaussian()
        \EndIf
        \EndFor
        \State $\mathcal{M} \leftarrow$ Store($\mathcal{G}_l$, $t_i$) \Comment{Update the Gaussian Identity Table}
        \State $i \leftarrow i + k$
        \EndWhile
        \State \Return $\mathcal{M}$
        \EndProcedure
    \end{algorithmic}
\end{algorithm}
\clearpage

\begin{table*}
    \centering
    \begin{tabular}{c  c c  c c  c c }
        \toprule
         \multirow{2}{*}{Method} & \multicolumn{2}{c}{split-cookie} & \multicolumn{2}{c}{chickchicken} & \multicolumn{2}{c}{torchocolate} \cr  
        \cmidrule(lr){2-3}\cmidrule(lr){4-5}\cmidrule(lr){6-7}
         ~ & IoU(\%) & Acc(\%) & IoU(\%) & Acc(\%) & IoU(\%) & Acc(\%)\\
         \midrule
         SAGA~\cite{cen2023segmentany3dgs} & 51.99 & 56.78 & 68.88 & 72.75 & 58.45 & 60.36\\
         Gaussian Grouping~\cite{ye2023gaussiangrouping} & 57.87 & 90.68 & 80.43 & 98.41 & 68.02 & 89.15 \\
         \midrule
         Ours w/o TFF (w/o Refinement)& 76.89 & 99.82 & 90.73 &\textbf{ 99.03} & 74.00 & \textbf{99.19} \\ 
         Ours w/ TFF (w/o Refinement) & 83.31 & \textbf{99.82} & 88.85 & 99.00 & 74.95 & 99.03 \\
         Ours w/ all& \textbf{88.92} & 99.73 & \textbf{92.57} & 98.53 & \textbf{88.92} & 98.34 \\
         \toprule
         \multirow{2}{*}{Method} & \multicolumn{2}{c}{espresso} & \multicolumn{2}{c}{keyboard} & \multicolumn{2}{c}{americano} \cr  
        \cmidrule(lr){2-3}\cmidrule(lr){4-5}\cmidrule(lr){6-7}
         ~ & IoU(\%) & Acc(\%) & IoU(\%) & Acc(\%) & IoU(\%) & Acc(\%)\\
         \midrule
         SAGA~\cite{cen2023segmentany3dgs} & 57.64 & 98.22 & 73.65 & 78.01 & 80.91 & 87.22  \\
         Gaussian Grouping~\cite{ye2023gaussiangrouping} & 62.36 & 79.66 & 74.77 & 93.32 & 73.71 & 98.08  \\
         \midrule
         Ours w/o TFF (w/o Refinement)& 72.39 & 99.88 & 86.72 & 99.44 & 80.80 & \textbf{99.99} \\ 
         Ours w/ TFF (w/o Refinement) & 74.11 & \textbf{99.89} & 82.87 & \textbf{99.52} & 82.48 & 99.98 \\
         Ours w/ all& \textbf{88.42} & 99.73 & \textbf{90.73} & 99.19 & \textbf{89.60} & 99.91  \\
         \bottomrule
    \end{tabular}
    \caption{Quantitative results on the HyperNeRF~\cite{park2021hypernerf} dataset. ``Ours w/o TF'' means directly  attaching an identity feature vector to each Gaussian.}
    \label{tab:hypernerf results}
\end{table*}

\begin{table*}
    \centering
    \resizebox{1.0\linewidth}{!}{
    \begin{tabular}{c c c c c c c}
        \toprule
         \multirow{2}{*}{Method} & \multicolumn{2}{c}{cut roasted beef} & \multicolumn{2}{c}{flame steak} & \multicolumn{2}{c}{cook spinach} \cr  
        \cmidrule(lr){2-3}\cmidrule(lr){4-5}\cmidrule(lr){6-7}
         ~ & IoU(\%) & Acc(\%) & IoU(\%) & Acc(\%) & IoU(\%) & Acc(\%)\\
         \midrule
         SAGA~\cite{cen2023segmentany3dgs} & 74.25 & 76.11 & 78.33 & 81.26 & 78.66 & 81.57 \\
         Gaussian Grouping~\cite{ye2023gaussiangrouping} & 84.86 & 93.74 & 88.94 & 99.87 & 85.60 & 99.06\\
         \midrule
         Ours w/o. Refinement & 85.05 & 99.97 & 76.14 & 99.82 & 82.01 & 99.97\\
         Ours & \textbf{94.09} & 99.88 & \textbf{92.63} & 99.40 & \textbf{92.33} & 99.75 \\
         \toprule
         \multirow{2}{*}{Method} & \multicolumn{2}{c}{flame salmon} & \multicolumn{2}{c}{coffee martini} & \multicolumn{2}{c}{sear steak} \cr  
        \cmidrule(lr){2-3}\cmidrule(lr){4-5}\cmidrule(lr){6-7}
         ~ & IoU(\%) & Acc(\%) & IoU(\%) & Acc(\%) & IoU(\%) & Acc(\%) \\
         \midrule
         SAGA~\cite{cen2023segmentany3dgs} & 75.39 & 78.21 & 77.97 & 82.40 & 72.96 & 89.79 \\
         Gaussian Grouping~\cite{ye2023gaussiangrouping} & 87.41 & 99.93 & 89.53 & 99.82 & 85.78 & 99.89 \\
         \midrule
         Ours w/o Refinement & 78.43 & 99.85 & 83.28 & 99.67 & 75.95 & 99.98 \\
         Ours w/ all& \textbf{93.10} & 99.77 & \textbf{93.07} & 99.80 & \textbf{92.91} & 99.93  \\
         \bottomrule
         
    \end{tabular}
    }
    \caption{Quantitative results on the Neu3D~\cite{li2022neural} dataset.}
    \label{tab:dynerf results}
\end{table*}

\clearpage

\clearpage
\clearpage

\subsection{More Limitations}
\label{more limitations}
\begin{figure}[ht]
    \centering
    \includegraphics[width=1\linewidth]{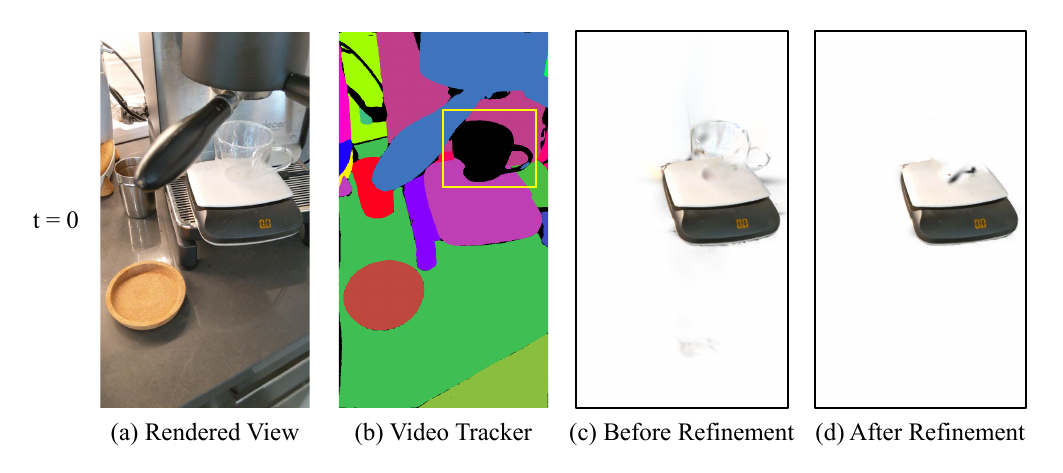}
    \vspace{-15pt}
    \caption{The failure case of 4D segmentation refinement.}
    \label{fig: limitation refine}
\end{figure}

\textbf{4D Segmentation Refinement.} In the 4D segmentation refinement, we utilize the segmentation mask of zero-shot video segmentation model~\cite{DEVA} as prior to remove ambiguous Gaussians at the boundary. However, \cite{DEVA} sometimes fail to generate correct segmentation masks, which leads to the failure of the 4D segmentation refinement. As illustrated in Fig.~\ref{fig: limitation refine}, since zero-shot video segmentation model labels the cup as void (the black region in the yellow box) at timestamp 0 by mistake, the whole cup is removed after refinement.

\begin{figure}[ht]
    \centering
    \includegraphics[width=1\linewidth]{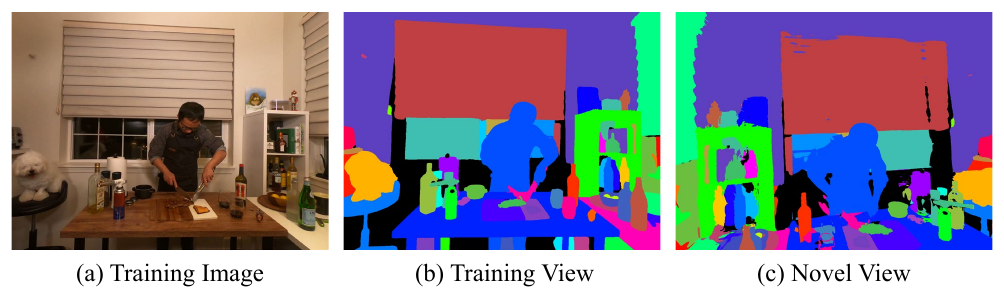}
    \vspace{-15pt}
    \caption{Novel view anything mask rendering results in the Neu3D~\cite{li2022neural} dataset.}
    \label{fig: limitation id collision}
\end{figure}

\textbf{Mask Identity Collision in Multi-view Settings.} In the Neu3D Dataset, we only use the image sequence under one camera view rather than mutiple views for training due to the Mask Identity Collision, in which video tracker~\cite{DEVA} assigns different ID to the same object in different video inputs. However, training with a single view may result in inferior anything masks when rendered in novel views, as shown in Fig.~\ref{fig: limitation id collision}. Moreover, 3D segmentation results without any refinement at each timestamp may also contain lots of noise due to heavy occlusion in a single view.

\subsection{Mask Annotation}
\label{mask annotation}
\begin{figure}[ht]
    \centering
    \includegraphics[width=1\linewidth]{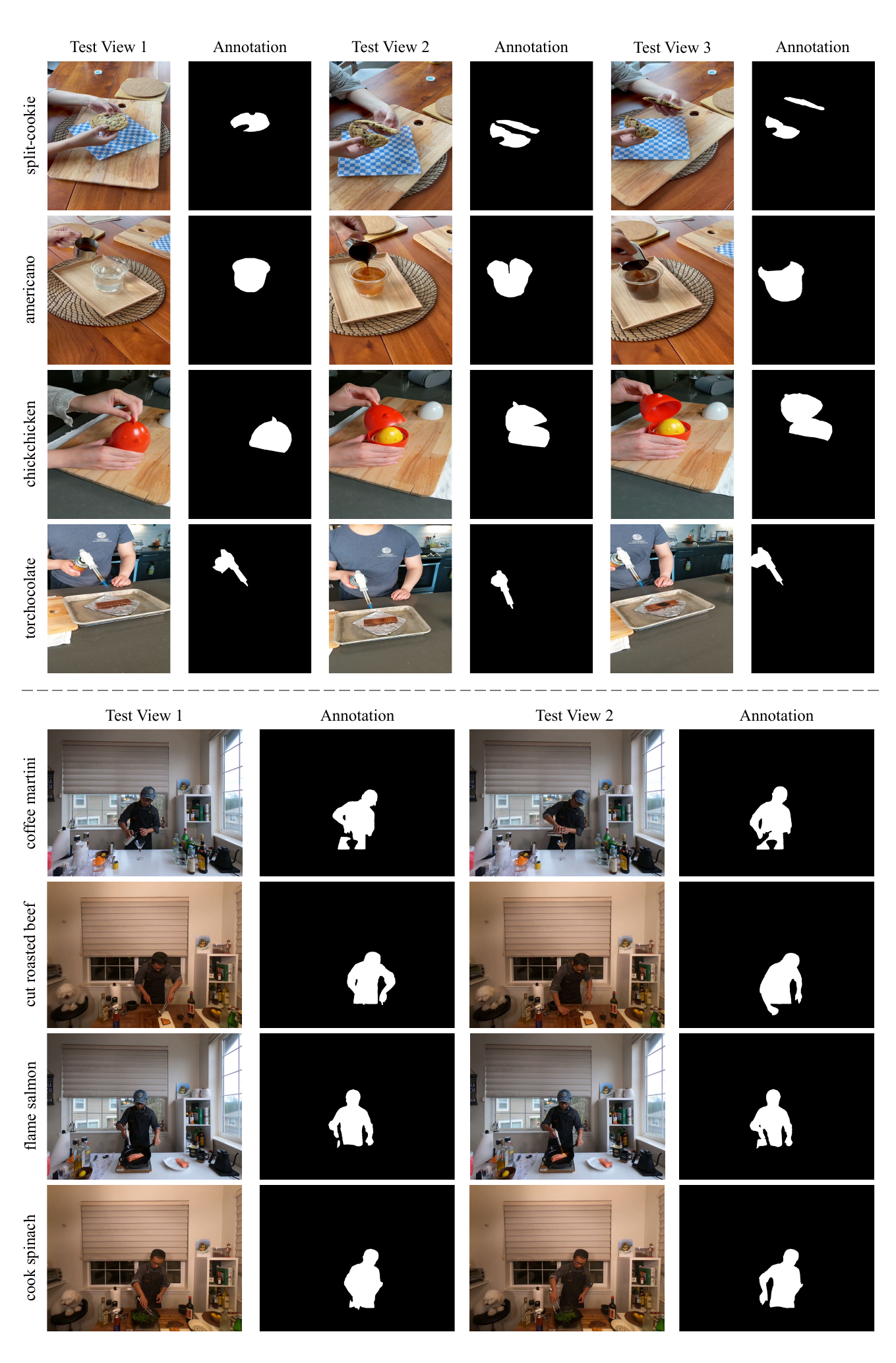}
    \vspace{-15pt}
    \caption{Visualizations of mask annotations. The top four rows are from the HyperNeRF~\cite{park2021hypernerf} dataset and the bottom four rows are from the Neu3D~\cite{li2022neural} dataset.}
    \label{fig: mask annotation}
\end{figure}

As there are no ground truth segmentation mask labels for HyperNerf~\cite{park2021hypernerf} and Neu3D~\cite{li2022neural} datasets, we manually annotate 6 scenes from HyperNerf and all scenes from Neu3D. We use the Roboflow platform and SAM~\cite{kirillov2023segmentanything} for interactive mask annotation. For each scene, we choose 20-30 frames in the test dataset and annotate object masks for measuring the segmentation accuracy. We provide some visualization examples of mask annotations in Fig.~\ref{fig: mask annotation}.

\subsection{More discussions}\label{more discussions}
\textbf{Training 4D-GS with semantic features.} \label{training 4dgs with semantic features}
There are several approaches~\cite{zhi2021place,ye2023gaussiangrouping} demonstrate that optimizing the scene with a semantic feature simultaneously may trigger the downgrade of rendering novel views. It is mainly because the object semantic feature may enforce the Gaussian included in $o$ have similar features. Therefore, our SA4D adopts two-stage training: training deformation field $\mathcal{F}$ and canonical 3D-GS $\mathcal{G}$ at first, then optimizing temporal identity feature field network by freezing the weight of $\mathcal{F}$ and $\mathcal{G}$.

\textbf{May SA4D help monocular dynamic scene reconstruction?}
Similar to GaussianObject~\cite{yang2024gaussianobject}, SA3D/SAGA is used to generate the object's mask, and the repair process is proposed to repair the 3D object. We believe that SA4D can also support the reconstruction in the monocular setups and leave it as future work.

\textbf{Social impacts.} SA4D shows the ability of composition with 4D Gaussians. The rendered image achieves photo-realistic performance. To mitigate the societal impact of highly realistic technology, several measures can be taken. Technologically, embedding watermarks or digital signatures and developing detection algorithms can ensure image authenticity. Legally, implementing regulations and strict enforcement can deter malicious activities. Educating the public on media literacy enhances their ability to discern real from fake images. Social media platforms must enforce rigorous content review and provide easy reporting mechanisms. Lastly, industry self-regulation and ethical training for practitioners can promote responsible 4D composition/editing practices. 
%

\end{document}